\pdfoutput=1
\documentclass[11pt]{article}

\usepackage[preprint]{acl}

\usepackage{times}
\usepackage{latexsym}
\usepackage{booktabs} % for professional tables
\usepackage[T1]{fontenc}
\usepackage[utf8]{inputenc}

\usepackage{microtype}
\usepackage{microtype}
\usepackage{graphicx}
\usepackage{subfigure}
\usepackage{booktabs} % for professional tables
\usepackage{amsmath}
\usepackage{amssymb}
\usepackage{mathtools}
\usepackage{amsthm}
\usepackage{fixmath}
\usepackage{amsmath}
\usepackage{amsfonts}
\usepackage{amssymb}
\usepackage{multirow}
\usepackage{booktabs}
\usepackage{multicol}
\usepackage{lscape} % Optional: For landscape tables
\usepackage{mathtools}
\usepackage{amssymb}% http://ctan.org/pkg/amssymb
\usepackage{pifont}% http://ctan.org/pkg/pifont
\newcommand{\cmark}{\ding{51}}%
\newcommand{\xmark}{\ding{55}}%

\usepackage{amsmath}
\usepackage{amssymb}
\usepackage{mathtools}
\usepackage{amsthm}
\usepackage{inconsolata}

\usepackage{graphicx}

\title{LaMDA: \underline{La}rge \underline{M}odel Fine-Tuning via Spectrally \underline{D}ecomposed Low-Dimensional \underline{A}daptation}

\author{
 \textbf{Seyedarmin Azizi\textsuperscript{1}\thanks{Correspondence author (\href{seyedarm@usc.edu}{seyedarm@usc.edu})}},
 \textbf{Souvik Kundu\textsuperscript{2}},
 \textbf{Massoud Pedram\textsuperscript{1}}
\\
\\
 \textsuperscript{1}University of Southern California, USA
 \textsuperscript{2}Intel Labs, San Diego, USA
}

\begin{document}
\maketitle
\begin{abstract}
Low-rank adaptation (LoRA) has become the default approach to fine-tune large language models (LLMs) due to its significant reduction in trainable parameters. However, trainable parameter demand for LoRA increases with increasing model embedding dimensions, leading to high compute costs. Additionally, its backward updates require storing high-dimensional intermediate activations and optimizer states, demanding high peak GPU memory. 
In this paper, we introduce \textit{LaMDA}, a novel approach to fine-tuning large language models, which leverages low-dimensional adaptation to achieve significant reductions in trainable parameters and peak GPU memory footprint. LaMDA freezes a first projection matrix (PMA) in the adaptation path while introducing a low-dimensional trainable square matrix, resulting in substantial reductions in trainable parameters and peak GPU memory usage. LaMDA gradually freezes a second projection matrix (PMB) during the early fine-tuning stages, reducing the compute cost associated with weight updates to enhance parameter efficiency further.
We also present an enhancement, LaMDA++, incorporating a ``lite-weight" adaptive rank allocation for the LoRA path via normalized spectrum analysis of pre-trained model weights. We evaluate LaMDA/LaMDA++ across various tasks, including natural language understanding with the GLUE benchmark, text summarization, natural language generation, and complex reasoning on different LLMs.
Results show that LaMDA matches or surpasses the performance of existing alternatives while requiring up to \textbf{17.7}$\times$ fewer parameter updates and up to \textbf{1.32}$\times$ lower peak GPU memory usage during fine-tuning. Code will be publicly available at \url{https://github.com/ArminAzizi98/LaMDA}.
\end{abstract}
\section{Introduction}
Large language models (LLMs) have demonstrated remarkable performance in addressing a variety of natural language processing (NLP) tasks due to their generalization ability upon training on large corpus of data \cite{DBLP:journals/corr/abs-2005-14165, DBLP:journals/corr/abs-2302-13971}. To fully harness the capabilities of LLMs, fine-tuning has become the standard approach to serve various downstream tasks. However, full fine-tuning of LLMs can be prohibitively costly, making fine-tuning at the edge hardly possible. For example, even the smaller variants of LLMs with 7B parameters may ask for $\sim60$ GB memory to perform full fine-tuning  \cite{DBLP:journals/corr/abs-2403-17919}. Additionally, such approach is prone to causing overfitting and catastrophic forgetting in the over-parameterized regime \cite{DBLP:journals/corr/abs-2308-08747, DBLP:journals/corr/abs-2401-04051}.
%Full fine-tuning involves training a substantial portion of the model parameters on the target tasks. Given the scale of state-of-the-art models, such as LLaMA2 \cite{DBLP:journals/corr/abs-2307-09288}, which contain billions of parameters, full fine-tuning requires large GPU memory. For instance, fine-tuning a model with 7 billion parameters requires approximately 60GB of memory \cite{DBLP:journals/corr/abs-2403-17919}. Moreover, the updates associated with training the parameters of large models incur massive computational costs and energy consumption \cite{DBLP:journals/corr/abs-2401-02038}. Additionally, full fine-tuning is prone to causing overfitting and catastrophic forgetting in the over-parameterized regime \cite{DBLP:journals/corr/abs-2308-08747, DBLP:journals/corr/abs-2401-04051}.
\begin{figure}[t]
    \centering
  \includegraphics[width=0.45\textwidth]{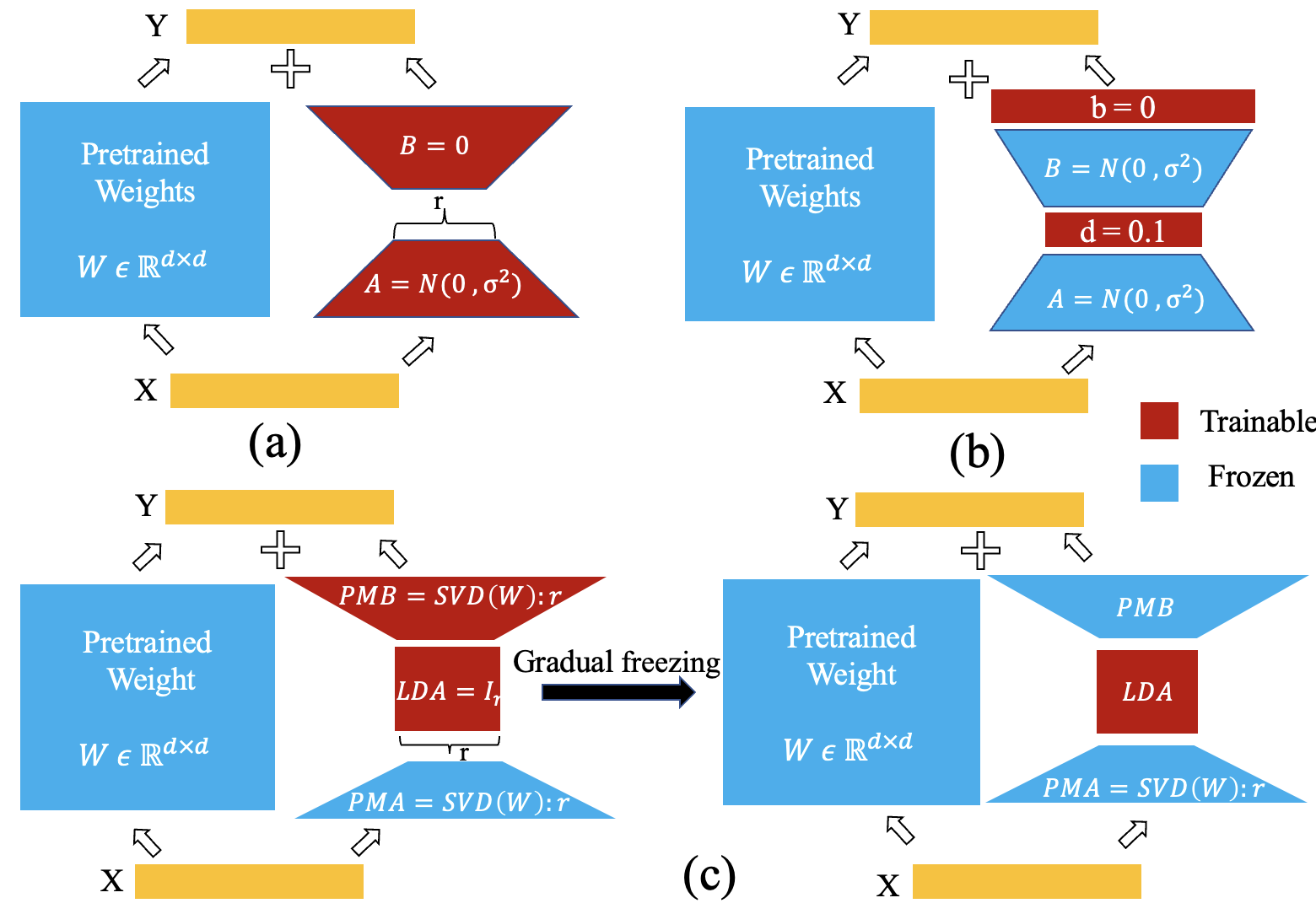}
  \caption{(a) LoRA \cite{DBLP:conf/iclr/HuSWALWWC22}. (b) VERA \cite{DBLP:journals/corr/abs-2310-11454}. (c) \textbf{LaMDA}. At the beginning, PMB is trainable and gradually freezes based on the singular values. After \(t_i\) iterations, PMB is completely frozen, and only the LDA is fine-tuned.}
  \label{fig:LaMDA}
  \vspace{-6mm}
\end{figure}

As a solution to these challenges, parameter-efficient fine-tuning (PEFT) techniques were proposed in which either a small portion of model parameters are updated, including the weight adapters \cite{DBLP:conf/icml/HoulsbyGJMLGAG19, DBLP:conf/emnlp/HuWLXLB0PL23}, or task-specific soft prompts are trained \cite{DBLP:conf/emnlp/LesterAC21}. Among these, low-rank adaptation (LoRA) \cite{DBLP:conf/iclr/HuSWALWWC22} has gained significant popularity.
It assumes that the changes in the pre-trained weight reside in a low-rank space and thus adds two trainable low-rank adapters, named the projection matrix \pmb{$A$} (PMA) and the projection matrix \pmb{$B$} (PMB) as \pmb{$BA$} in parallel to the frozen main path of the model weight \pmb{$W$} (refer to Fig. \ref{fig:LaMDA}(a)). LoRA fine-tuning can reduce GPU memory demand and trainable parameter-count since it only fine-tunes the \pmb{$BA$} which is much smaller in parameter count compared to \pmb{$W$}.
% \begin{figure}[t]
%   \includegraphics[width=0.8\columnwidth]{LoRA.png}
%   \caption{Low-rank adaptation (LoRA)}
%   \label{fig:Lora}
% \end{figure}
However, the number of trainable-parameters in LoRA may still be potentially larger than the low intrinsic dimension of a pre-trained LLM \cite{DBLP:conf/acl/AghajanyanGZ20}. Moreover, as evident in Fig \ref{fig:LaMDA}(a), the input activations \pmb{$X$} that must be stored for backpropagation reside in a \(d\)-dimensional space, where \(d\) denotes the embedding dimension of the model. Subsequently, the activation's GPU memory increases linearly with the embedding size, and LoRA does not provide any merit in activation memory saving. Notably, few contemporary works \cite{DBLP:journals/corr/abs-2403-13269, DBLP:journals/corr/abs-2310-11454} have presented solutions of \pmb{$BA$} freezing. However, they still suffer from increased activation storage and often demand high rank to compensate for the accuracy drop due to freezing \cite{DBLP:journals/corr/abs-2310-11454}. 

To address these issues, in this work we present, $\textbf{La}$rge $\textbf{M}$odel Fine-tuning via Spectrally $\textbf{D}$ecomposed Low-Dimensional $\textbf{A}$daptation (LaMDA). LaMDA as demonstrated in Fig. \ref{fig:LaMDA}(c) employs a trainable \textbf{low-dimensional adapter} (LDA), which is a square matrix in the \(r\)-dimensional space. We keep the PMA frozen throughout the fine-tuning, while allow PMB to freeze gradually only after few epochs based on relative magnitude of the singular values. We only keep the LDA trainable throughout. This allows the trainable parameters to be independent of \(d\)\ and the activations that are saved for backward pass remain in the \(r\)-dimensional space (\(r \ll d\)). Thus LaMDA can significantly reduce the trainable parameter, activation, and optimizer state memory. We summarize our contributions as follows:
%This decision significantly decreases the trainable parameters and the memory required for storing activations during back-propagation. The reason for memory saving is the observation that the activations that are saved for backward pass are in the \(r\)-dimensional space (\(r \ll d\)). To sustain performance levels comparable to LoRA for more complex tasks while benefiting from these reductions, LaMDA gradually freezes the PMB matrix in the early stages of fine-tuning. PMA and PMB are initialized using the left and right singular vectors corresponding to the most significant (largest) singular values of the pre-trained weight \(W\), respectively. PMB is then progressively frozen based on the relative magnitudes of the singular values. This procedure keeps the effective number of trainable parameters very low.
\begin{itemize}
    \item We introduce LaMDA, a novel framework to fine-tune LLMs that significantly reduces both parameter count and activation memory, resulting in lower computational costs and GPU memory usage.
    \item We then present LaMDA++, a novel enhancement of LaMDA that uses adaptive rank across different layers to fine-tune the model. Precisely, we use the pre-trained weight tensors to present a ``lite-weight" normalized \textbf{energy-score}\footnote{Energy-score of a matrix is defined as the summation of the square of its singular values.} based layer ranking to adaptively assign rank to the LDA of each layer allowing more optimal distribution of trainable parameters. Table \ref{tab:comp_app} compares the different PEFT methods and their benefits and limitations in the context of different memory footprint and adaptive rank allocation policy.
    %\item We compare LaMDA to state-of-the-art fine-tuning methods, including LoRA \cite{DBLP:conf/iclr/HuSWALWWC22}, VERA \cite{DBLP:journals/corr/abs-2310-11454}, LoRA-FA \cite{DBLP:journals/corr/abs-2308-03303}, and AFLoRA \cite{DBLP:journals/corr/abs-2403-13269}, in terms of computation cost, number of trainable parameters, and peak GPU memory usage. Our results demonstrate that LaMDA provides a superior trade-off without incurring additional inference costs.
    %alternately known as the square of their Frobenius norm.}  
    \item We evaluate the performance of LaMDA and LaMDA++ fine-tuning on encoder-only (DeBERTa-V3 \cite{he2021debertav3}), encoder-decoder (BART-Large \cite{DBLP:conf/acl/LewisLGGMLSZ20}), and decoder-only (LLaMA2-7B \cite{DBLP:journals/corr/abs-2307-09288}) models across various tasks including the GLUE benchmark for natural language understanding,  text summarization, and  complex reasoning. Our experiments show that LaMDA fine-tuned models consistently yield similar or improved performance with up to \textbf{17.7}$\times$ fewer trainable parameters, at reduced activation storage while providing peak GPU memory saving of up to \textbf{1.32}$\times$.
\end{itemize}

\begin{table}[!t]
    \centering
    \caption{Comparison of different important metrics associated to different fine-tuning techniques.}
    \label{tab:comp_app}
    \vspace{-3mm}
    \resizebox{0.90\columnwidth}{!}{%
    \begin{tabular}{lccccc}
    \toprule
    {Method} & \multicolumn{3}{c}{Memory} & Adaptive\\
    \cline{2-4}
    {}    & Optimizer & Gradient & Activation & {rank} \\
    \midrule
    Full FT & \xmark  & \xmark & \xmark & \xmark   \\
    LoRA  & \cmark  & \cmark & \xmark & \xmark   \\
    AdaLoRA  & \cmark  & \cmark & \xmark & \cmark   \\
    LoRA-FA       & \cmark  & \cmark & \cmark & \xmark  \\
    LISA       & \cmark  & \cmark & \xmark  & \xmark  \\
    VERA       & \cmark  & \cmark \cmark & \xmark & \xmark  \\
    AFLoRA       & \cmark  & \cmark \cmark & \xmark & \xmark  \\
    \textbf{LaMDA} (\textbf{Ours}) & \cmark & \cmark \cmark & \cmark & \xmark \\
    \textbf{LaMDA++} (\textbf{Ours}) & \cmark & \cmark \cmark & \cmark & \cmark \cmark \\
    \bottomrule
    \end{tabular}
    }
    \begin{flushleft}

\end{flushleft}
    % \end{sc}
% \end{center}
\vskip -0.2in
\end{table}

\section{Background}
\textbf{Transformer-based models.} Each module of an  $L$ layer transformer model \cite{vaswani2017attention} usually consists of two sub-blocks: the multi-head self-attention (MHSA) sub-block and the feed-forward network (FFN). Each MHSA takes input token embedding \(\pmb{X} \in \mathbb{R}^{n\times d}\) and applies the following:
\begin{equation}
    \pmb{Q}^{(i)} = \pmb{X}\pmb{W}^{(i)}_Q, \space \pmb{K}^{(i)} = \pmb{X}\pmb{W}^{(i)}_K, \space \pmb{V}^{(i)} = \pmb{X}\pmb{W}^{(i)}_V
\end{equation}
\begin{equation}
    \pmb{H}^{(i)} = [\mathrm{Softmax}(\pmb{Q}^{(i)}\pmb{K}^{(i)T}/\sqrt{d_h})\pmb{V}^{(i)}]
\end{equation}
\begin{equation}
    \mathrm{MHSA}(\pmb{X}) = \text{Concat}[\pmb{H}^{(1)},.., \pmb{H}^{(i)},.., \pmb{H}^{(h)}]\pmb{W}_o
\end{equation}
where \(\pmb{W}_O\) \(\in \mathbb{R}^{d\times d}\), \(\pmb{W}_{Q,K,V}\) (all \(\in \mathbb{R}^{d\times d_h}\)) are output projection, query, key, and value matrices with $d_h$ as model embedding dimension per head. The FFN includes two linear transformation layers ($\pmb{W_{FFN1}}$ and $\pmb{W_{FFN2}}$) with a non-linear activation function \(\sigma\) in the middle: \(\mathrm{FFN}(\pmb{X}) =\sigma(\pmb{XW_{FFN1}})\pmb{W_{FFN2}}\). With MHSA and FFN as the two sub-blocks, the output of the transformer block is computed as: 
\begin{equation}
    \pmb{X}^\prime = \mathrm{LayerNorm}(\pmb{X}+\mathrm{MHSA}(\pmb{X}))
\end{equation}
\begin{equation}
    \pmb{Y} = \mathrm{LayerNorm}(\pmb{X}^\prime+\mathrm{FFN}(\pmb{X}^\prime))
\end{equation}
LayerNorm is the layer normalization module, and Y is the output of the transformer block. 
\newline
\noindent
\textbf{Low-rank adaptation}. LoRA adds a trainable adaptation path (through the \pmb{$A$} and \pmb{$B$} matrix) parallel to the frozen main path of the respective module (i.e., \pmb{$W$}). This results in a significant reduction in the number of trainable parameters, the optimizer's state memory, and the required gradient memory:
\begin{equation}
    \label{lora}
    \pmb{Y} = \pmb{XW} + \alpha \pmb{XAB}.
\end{equation}
$\alpha$ serves as a fine-tuning hyper-parameter. 

\begin{figure}[t]
  \includegraphics[width=0.9\columnwidth]{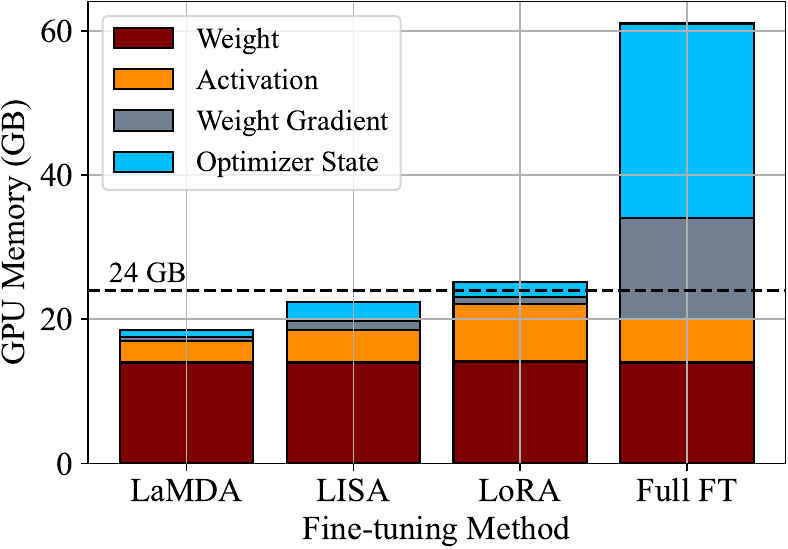}
  \caption{GPU memory usage of LLaMA2-7B on different fine-tuning methods including ours (LaMDA).}
  \vspace{-6mm}
  \label{fig:LaMDA-memory}
\end{figure}

Further numerous variants of LoRA have been introduced to enhance its performance. Earlier works \cite{zhang2023adaptive, DBLP:journals/corr/abs-2403-17919} explored different per-layer rank allocation and layer-importance sampling strategies to better utilize the fine-tuning budget across the model layers. The approach learns the adapters' ranks dynamically by analyzing the singular values of the adapters, allowing for more effective utilization of the fine-tuning budget.
More recently, \cite{DBLP:journals/corr/abs-2308-03303} addresses GPU memory savings by freezing the PMA matrix in the adaptation path, thereby reducing the size of the stored activations during fine-tuning. However, this method still requires fine-tuning \(d \times r\) parameters per linear layer and compromises accuracy. VERA \cite{DBLP:journals/corr/abs-2310-11454} takes a different approach by randomly initializing and freezing PMA and PMB with a large \(r\) dimension, focusing on fine-tuning two feature transformation vectors instead (Fig. \ref{fig:LaMDA}). While this method reduces parameter count, add significant compute and activation memory overhead. 
To address VERA's computational inefficiencies, AFLoRA \cite{DBLP:journals/corr/abs-2403-13269} was proposed. However, it still suffers from high activation storage overhead.

 LaMDA, on the contrary, offers two key benefits: 1) it significantly reduces trainable parameter, activation, and optimizer storage to enhances memory savings compared to LoRA; 2)  it greatly reduces computational cost in the forward pass during fine-tuning. Table \ref{tab:comp_app} compares various state-of-the-art (SOTA) fine-tuning methods regarding their memory requirement and rank adaptation. Notice that \textbf{LaMDA(++) is the \textbf{only method} that can simultaneously reduce gradient, optimizer, and activation memory while also yielding adaptive ranks based on a notion of layers' energy-score}.

\section{Methodology}
This section provides a detailed explanation of LaMDA and LaMDA++ as novel parameter-efficient fine-tuning methods.

\subsection{Low-Dimensional Adapter (LDA)}
\label{sec:lda}
One of the essential components of the LaMDA method is a square \(r\)-dimensional matrix \pmb{$S$}, as depicted in Figure \ref{fig:LaMDA}(c). Integrating this module into the adapter path yields the following formulation:
\begin{equation}
    \label{lamda}
    \pmb{Y} = \pmb{XW} + \alpha \pmb{XASB},
\end{equation}
where \(\pmb{A} \in \mathbb{R}^{d \times r}\), \(\pmb{S} \in \mathbb{R}^{r \times r}\), and \(\pmb{B} \in \mathbb{R}^{r \times d}\) represent PMA, LDA, and PMB, respectively. By freezing \(\pmb{A}\) and \(\pmb{B}\) and keeping \(\pmb{S}\) trainable, we significantly reduce the number of trainable parameters, which is reduced to \(r^2\ \ll 2 \times d \times r\) of LoRA, and is independent of the increasing model \(d\). This reduction in the number of trainable parameters offers a two-fold advantage. Firstly, it effectively constrains the parameter count, thereby reducing the risk of overfitting and enhancing the model's generalization capabilities. This is particularly advantageous considering that \(2 \times d \times r \times L\) potentially exceeds the intrinsic dimension of large language models \cite{DBLP:conf/acl/AghajanyanGZ20}. Secondly, fine-tuning fewer parameters requires less computation in the backward pass as fewer gradient-based updates and optimizer states computations happen accordingly. This alleviates the overall computational and optimizer storage overhead of fine-tuning. 
Additionally, employing the low-dimensional adapter while freezing \pmb{$A$} reduces activation memory usage during fine-tuning. Assuming a fine-tuning batch size of \(b\) and an input sequence length of \(n\), in LoRA, the input \(X\) in Equation \ref{lora} is represented as a \(\pmb{B} \times n \times d\) tensor. This tensor must be stored in GPU memory, as it is essential for computing the gradient for the trainable matrix \(\pmb{A}\). Consequently, the required GPU memory for storing the activations is a function of the embedding dimension \(d\). However, when utilizing the low-dimensional adapter \(\pmb{S}\) in Equation \ref{lamda}, and since \(\pmb{A}\) is not being trained, the required activation in the backward pass is of dimension \(\pmb{B} \times n \times r\). This leads to a significant reduction in the number of stored elements and GPU memory usage.

Figure \ref{fig:LaMDA-memory} reports the peak GPU memory usage of different fine-tuning methods for the LLaMA2-7B model with a batch size one. As the figure shows, compared to LoRA, most GPU memory saving is achieved by the required activation memory reduction. Furthermore, LaMDA surpasses the current SOTA fine-tuning approach of LISA \cite{pan2024lisa}. Having established the benefits of our low-dimensional adapter, we delve into the details of the LaMDA fine-tuning process in the next section.

\subsection{LaMDA}
\label{sec:lamda}

Building upon the idea of the low-dimensional adapter, we now disclose the LaMDA in detail. Considering Figure \ref{fig:LaMDA}(c), a natural implementation of the idea of the low-dimensional adapter would be to keep the \(\pmb{A}\) and \(\pmb{B}\) frozen and train the matrix S until convergence. This achieves the benefits discussed in section \ref{sec:lda}. One critical issue will be the initialization of the fixed adapters PMA and PMB. VERA \cite{DBLP:journals/corr/abs-2310-11454} kept them frozen by initializing via Kaiming normal distribution. Despite frozen \pmb{$BA$}, the downside was that it required the rank \(r\) to potentially converge to good accuracy, thus costing substantial compute for the forward and the backward pass. We on the contrary, propose to initialize PMA and PMB with the singular vectors (SVs) corresponding to the most significant singular values of the pre-trained weight. This is accomplished by applying singular value decomposition (SVD) to the pre-trained weight and extracting its \textbf{spectrum}, then initializing \(\pmb{A}\) and \(\pmb{B}\) with the SVs:
%Meanwhile, PISSA \cite{meng2024pissa} has shown that fine-tuning singular vectors offers benefits over random initialization of adapters. Sharing a high-level spirit with \cite{meng2024pissa}, we initialize PMA and PMB with the singular vectors (SV) corresponding to the most significant singular values of the pre-trained weight. This is accomplished by applying singular value decomposition (SVD) to the pretrained weight and extracting its \textbf{spectrum}, then initializing \(\pmb{A}\) and \(\pmb{B}\) with the SVs:
\begin{equation}
    \label{SVD}
    \pmb{U},\pmb{\Sigma},\pmb{V} = SVD(\pmb{W})
\end{equation}
\begin{equation}
    \label{SVD1}
    \pmb{A} = \pmb{U}[:,:r]\pmb{\Sigma} [:r,:r],\quad \pmb{B}= \pmb{V}[:,:r]^\mathrm{T}. 
\end{equation}
Since \(\pmb{B}\) forms a basis for \(\mathbb{R}^r\), learning matrix \(\pmb{S}\) in Equation \ref{lamda} can be interpreted as learning a \textbf{basis change matrix}. Previous studies \cite{DBLP:conf/iclr/HuSWALWWC22,DBLP:journals/corr/abs-2310-08659} have emphasized the importance of ensuring that the combined effect of the main path and the adapters approximates the pre-trained weights at the onset of fine-tuning. Accordingly, based on Equations \ref{SVD} and \ref{SVD1}, we initialize \(\pmb{S}\) (LDA) with the identity matrix \(I_r\) and set the main path with the last \( d-r \) components of  spectrum of \(\pmb{W}\). We note, a contemporary work \cite{meng2024pissa} has suggested similar initialization of the adapters. However, our approach largely differ  in primary objective, as we intend to find suitable initialization to freeze by allowing the LDA to learn. \cite{meng2024pissa}, on the contrary, focuses primarily on the impact of adapter initialization and does not yield any memory or compute saving compared to that with LoRA.

Having initialized all the parameters in equation \ref{lamda}, we perform fine-tuning by keeping the PMA always frozen and LDA always learnable. For the PMB, we present a gradual freezing strategy, to be discussed next. We hypothesize that having only a trainable LDA for simpler tasks (e.g. GLUE benchmark) would be sufficient potentially due to much lower intrinsic dimensions of the pre-trained weights, thus not necessitating any need of high trainable parameters.
However, for relatively complex tasks, like summarizing, complex reasoning, we believe intrinsic dimensionality of the weight may not be very low. To tackle this challenge, we adapt a gradual freezing strategy of the PMB allowing the fine-tuned model to perform better while keeping all the benefits of LaMDA.
Further analysis on the relation of task difficulty to model intrinsic low-dimensionality may be an interesting future research.

To enhance LaMDA's expressiveness while maintaining the benefits of having low parameter count and minimal activation memory, we introduce the mechanism of \textbf{gradual freezing} of PMB, as illustrated in Figure \ref{fig:LaMDA}(c). The concept involves keeping PMB trainable during the initial iterations of fine-tuning and then progressively freeze PMB row by row. Previous work by \cite{DBLP:journals/corr/abs-2403-13269} has suggested gradual freezing based on fixing the scores computed during fine-tuning. In contrast, we employ a simpler heuristic to circumvent the additional computational and memory overhead associated with calculating and storing these scores. As argued in \cite{meng2024pissa}, learning the SVs corresponding to the most significant singular values is the most effective approach for parameter-efficient fine-tuning. Consequently, a reasonable criterion is to freeze the rows of \(\pmb{B}\) sequentially from the last row to the first, given that the first row represents the highest-energy component of the spectrum of PMB. So, we propose a linear schedule for the number of trainable rows in PMB as below:
\begin{equation*}
    r(t)=
    \begin{cases}
        \mathrm{int}(r -\frac{t}{t_i}), & \text{$t \leq t_i$} \\
        0 & \text{ $t \geq t_i$}
    \end{cases}
\end{equation*}
where \(t_i\) is set to be 30\% of the total iterations in our experiments. Since \(\pmb{B}\) is an \(r \times d\) matrix, the input activation that needs to be stored for backpropagation is again in the \(r-\)dimensional space, so the memory-saving arguments still hold. In section \ref{sec:ablation}, we do an ablation study on the correct order of freezing the rows of PMB. 

\begin{figure}[t]
  \includegraphics[width=0.9\columnwidth]{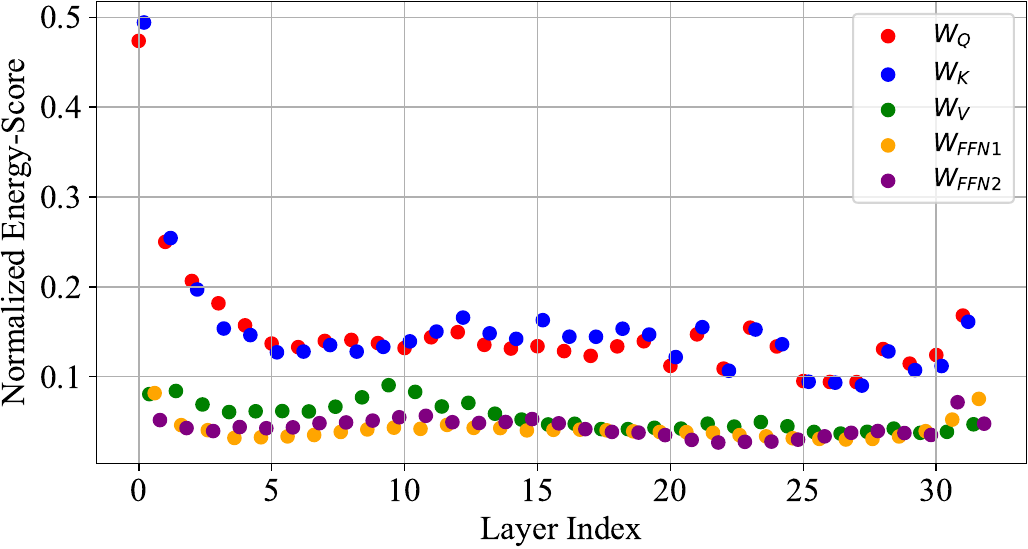}
  \caption{Layer-wise energy-score of the first 32 ranks of each linear module, normalized over the total energy-score of the same module, evaluated on a pre-trained LLaMA2-7B.}
  \label{fig:energy}
\end{figure}

% \begin{figure*}[t]
%   \includegraphics[width=0.48\linewidth]{energy.pdf} \hfill
%   \includegraphics[width=0.48\linewidth]{new_energy.pdf}
%   \caption {A minimal working example to demonstrate how to place
%     two images side-by-side.}
% \end{figure*}

\subsection{LaMDA++}
\label{sec:lamda++}
This section presents LaMDA++, an enhanced version of LaMDA that incorporates the option of varying ranks across different network layers. Previous works on adaptive rank for LLMs \cite{zhang2023adaptive} have introduced multiple hyperparameters, that can lead to increased training time. Furthermore, changing the rank of the matrices dynamically could result in more complex training. Conversely, We rely on a ``lite-weight" static analysis for fine-tuning with adaptive rank. In specific,  we analyze the normalized energy-score of the pre-trained model weights to simplify implementation and adoption. 

To motivate this approach, Figure \ref{fig:energy} reports the normalized energy-score of the first 32 singular vectors ($E^l_{r}$ with $r=32$) for each trainable linear module \(l\) of a LLaMA2-7B across all layers. The normalization factor is the total energy-score computed over all the singular vectors ($E^l_T$) of the corresponding module $l$. As the figure indicates, some modules capture significantly higher normalized energy-score than others when applying SVD to the weights. This observation suggests that, to achieve a similar normalized energy-score across all layers, the \(\pmb{W_Q}\) and \(\pmb{W_K}\) modules may require a lower rank. In contrast, the \(\pmb{W_{FFN1}}\), \(\pmb{W_{FFN2}}\), and \(\pmb{W_V}\) modules might necessitate a higher rank to reach an equivalent level of normalized energy-score. This can be of great importance, as previous works \cite{DBLP:conf/emnlp/HuWLXLB0PL23} have shown that all linear layers of the LLMs (including the attention weights) are essential to be fine-tuned. %We focus on the energy level of the adapters because the components of the pre-trained weight's spectrum with the highest energy level are the \textbf{best candidates} to fine-tune \cite{meng2024pissa}. 

To implement this heuristic while maintaining the same number of trainable parameters, LaMDA++ employs a pre-processing step to select the ranks of each LoRA path. Intuitively, ranks should be reduced from the budget of layers less affected by rank reduction and reallocated to layers that capture the least normalized energy-scores. Firstly, we define a rank budget set \(R_S\), containing $S$ potential candidate ranks, \(R_S = \{r_1, ..., r_S\}\), with $r_1 < r_2 < r_S$, to be selected for a LoRA path. %We use \(R_S = \{16, 24, 32, 40, 48\}\). 
We ensure that the summation of all the different ranks selected for the layers gets averaged to the target rank $r_T$. Additionally, for a module at layer $l$, LaMDA++ assigns a candidacy score \(\nu_l\) as,
\begin{equation}
    \label{eq:lambda++}
    \nu_l = \frac{{E^l_{r_S}}-{E^l_{r_1}}}{E^l_{r_T}}
\end{equation}
%Here, \(\Sigma\) is the matrix of singular values obtained from SVD, and \(r_{min}\) and \(r_{max}\) represent the minimum and maximum possible ranks in the set \(R\), respectively. 
LaMDA++ then sorts the linear modules based on the ascending order of \(\nu_l\). The initial elements of this sorted array are the layers that potentially require higher ranks to yield better energy-scores. In contrast, the later elements can potentially sacrifice rank reduction without losing significant energy. 
Based on this ranking, and to maintain simplicity, LaMDA++ assigns \(r_{S}\) to the first \(\frac{1}{S}\)th quantile of the sorted array, \(r_{S-1}\) to the second \(\frac{1}{S}\)th quantile, and so on. This heuristic approach favors allocating higher rank to modules that would benefit most and lower rank to modules that would suffer the least from rank reduction.

% Use \verb|\appendix| before any appendix section to switch the section numbering over to letters. See Appendix~\ref{sec:appendix} for an example.

% \section{Bib\TeX{} Files}
% \label{sec:bibtex}

% Unicode cannot be used in Bib\TeX{} entries, and some ways of typing special characters can disrupt Bib\TeX's alphabetization. The recommended way of typing special characters is shown in Table~\ref{tab:accents}.

% Please ensure that Bib\TeX{} records contain DOIs or URLs when possible, and for all the ACL materials that you reference.
% Use the \verb|doi| field for DOIs and the \verb|url| field for URLs.
% If a Bib\TeX{} entry has a URL or DOI field, the paper title in the references section will appear as a hyperlink to the paper, using the hyperref \LaTeX{} package.
\section{Experiments}

This section evaluates LaMDA and LaMDA++ on NLU, NLG, and reasoning tasks.

\subsection{Experimental Setup}

\begin{table*}[ht]
\centering
\caption{Comparison of different fine-tuning methods for DeBERTa-V3 on GLUE benchmark.}
\label{tab:glue}
\resizebox{\textwidth}{!}{%
\begin{tabular}{@{}lcccccccccccc@{}}
\toprule
\textbf{Method} & \textbf{\#Params.} & \textbf{CoLA $\uparrow$} & \textbf{SST-2 $\uparrow$} & \textbf{MRPC $\uparrow$} & \textbf{QNLI $\uparrow$} & \textbf{STS-B $\uparrow$} & \textbf{RTE $\uparrow$} & \textbf{MNLI $\uparrow$} & \textbf{QQP $\uparrow$} & \textbf{Avg. $\uparrow$} \\ \midrule
FFT & 184M & 69.21 & 95.64 & 89.22 & 93.78 & 91.59 & 82.49 & 89.98 & 92.05/89.31 & 87.82 \\
LoRA (\(r\) = 8) & 1.33M & 69.73 & 95.57 & 89.71 & 93.76 & \textbf{91.86} & 85.32 & 90.46 & 91.95/89.26 & 88.38 \\
AdaLoRA & 1.27M & 70.86 & 95.95 & 90.22 & 94.28 & 91.39 & 87.36 & 90.30 & 92.13/88.41 & 88.83 \\
% SoRA (r = 4) & 0.47M & 71.05 & 95.57 & 90.20 & 93.92 & 91.76 & 86.04 & 90.43 & 92.06/89.44 & 88.71 \\
VERA & 0.16M & 70.74 & 95.18 & 90.93 & 93.58 & 91.08 & 87.36 & 90.22 & 90.69/87.63 & 88.53 \\
AFLoRA (\(r\) = 4) & 0.14M & 72.01 & 96.22 & \textbf{91.91} & \textbf{94.42} & 91.84 & \textbf{88.09} & 90.17 & 90.81/87.77 & 89.23 \\ 

LaMDA (\(r\) = 32) & \textbf{0.075M} & 71.60 & 95.70 & 90.44 & 93.72 & 91.30 & 87.50 & 90.05 & 90.70/87.70 & 88.87 \\ 

LaMDA++ (\(r_T\) = 32) & \textbf{0.078M} & \textbf{72.12} & \textbf{96.25} & 91.65 & 94.30 & 91.55 & 88.01 & \textbf{90.56} & 90.80/87.75 & \textbf{89.28} \\ \bottomrule

\end{tabular}%
}
\end{table*}
Our experiments encompass a broad range of models and datasets. For NLU, we utilize DeBERTa-V3 \cite{DBLP:conf/iclr/HeGC23} and conduct evaluations on the GLUE benchmark \cite{DBLP:conf/iclr/WangSMHLB19}. For NLG, we employ BART-large \cite{DBLP:conf/acl/LewisLGGMLSZ20} and assess performance on the XSUM \cite{DBLP:conf/emnlp/NarayanCL18} and CNN/DailyMail \cite{DBLP:conf/nips/HermannKGEKSB15} datasets. Additionally, we evaluate the LLaMA2 series \cite{DBLP:journals/corr/abs-2307-09288} on GSM8K \cite{DBLP:journals/corr/abs-2110-14168}, Wikitext-2 \cite{DBLP:conf/iclr/MerityX0S17}, and a collection of commonsense reasoning datasets.
\newline
Following prior works on LoRA variants \cite{DBLP:conf/iclr/HuSWALWWC22, zhang2023adaptive, meng2024pissa}, we freeze the main path of the model while treating the LoRA path according to the LaMDA methodology. LaMDA is applied to the MHSA and FFN blocks of all models, encompassing the \(\pmb{W_Q}\), \(\pmb{W_K}\), \(\pmb{W_V}\), \(\pmb{W_{FFN1}}\), and \(\pmb{W_{FFN2}}\) linear modules. As baselines, we compare LaMDA with full fine-tuning, LoRA, LoRA-FA \cite{DBLP:journals/corr/abs-2308-03303}, AFLoRA \cite{DBLP:journals/corr/abs-2403-13269}, and VERA \cite{DBLP:journals/corr/abs-2310-11454}. Our implementation of LaMDA is based on HuggingFace's Transformers library \cite{DBLP:journals/corr/abs-1910-03771}, and all experiments are conducted on a single NVIDIA A6000 GPU.

\vspace{-0.15mm}
\subsection{Encoder-only Model: DeBERTa-V3}
We fine-tuned DeBERTa-V3 \cite{DBLP:conf/iclr/HeGC23} using LaMDA and LaMDA++ on the GLUE NLU benchmark. For LaMDA, the rank of the adapter path is set to 32, and in LaMDA++, the target rank \(r_T\) is set to 32 as well. For LaMDA++, the set of potential candidate ranks is \(R_S = \{16, 24, 32, 40, 48\}\). For further details on experimental hyperparameters please refer to Appendix \ref{sec:appendix_train}. Table \ref{tab:glue} presents the performance and the number of trainable parameters for LaMDA, LaMDA++, and SOTA PEFT methods. As shown in the Table, LaMDA achieves performance close to LoRA with a 17.7\(\times\) reduction in the number of trainable parameters. Similarly, LaMDA achieves reductions of 17\(\times\), 2.1\(\times\), and 1.8\(\times\) compared to AdaLoRA, VERA, and AFLoRA, respectively. Furthermore, LaMDA++ achieves SOTA performance with only a negligible increase in the parameter count. 
Please note, here we trained the LDA only while keeping the PMA, PMB frozen throughout the fine-tuning period.

\subsection{Encoder-Decoder Model: BART-large}
\begin{figure}[t]
  \includegraphics[width=0.82\columnwidth]{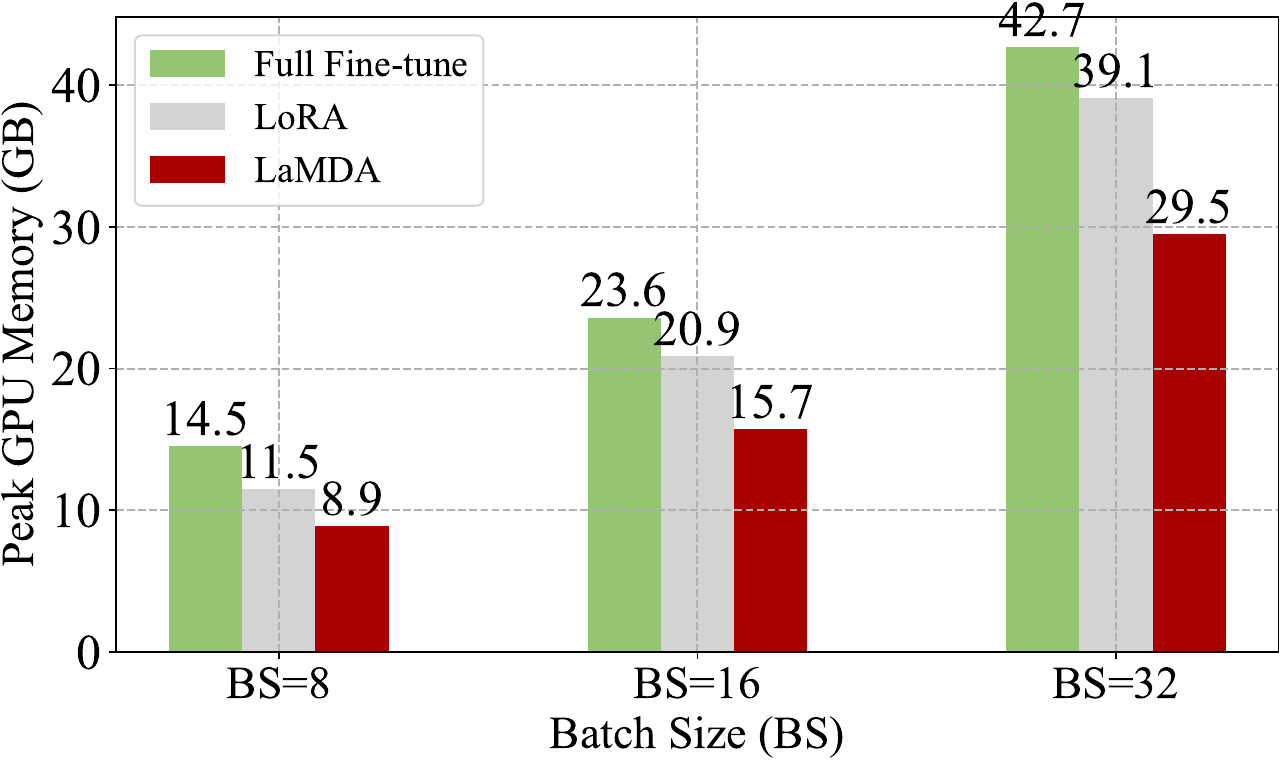}
  \caption{Peak GPU memory usage during fine-tuning BART-large on XSUM dataset.}
  \label{fig:chart_gpu}
  \vspace{-4mm}
\end{figure}

For the text summarization tasks, we utilize the BART-large model \cite{DBLP:conf/acl/LewisLGGMLSZ20} and fine-tune it on the XSUM \cite{DBLP:conf/emnlp/NarayanCL18} and CNN/DailyMail \cite{DBLP:conf/nips/HermannKGEKSB15} datasets using LaMDA. The selected rank and the set \(R_S\) are the same as those used for DeBERTa-V3. The low-rank path is added parallel to the math path of the MHSA and FFN blocks of the encoder and decoder across all model layers. As mentioned in section \ref{sec:lamda}, here we freeze PMA, keep LDA trainable, and gradually freeze PMB. The hyperparameter \(t_i\) is set to be 30\% of the total training iterations. For evaluation, we report the ROUGE-1, ROUGE-2, and ROUGE-L scores (R1/2/L) \cite{lin-2004-rouge}. Table \ref{tab:bart-large} showcases the number of trainable parameters and the performance of LaMDA and LaMDA++. Compared to LoRA, LaMDA achieves comparable performance while requiring 10\(\times\) fewer parameter updates. LaMDA++ surpasses LoRA on the XSUM dataset and performs similarly to it on CNN/DailyMail. The hyperparameters used for fine-tuning are provided in appendix \ref{sec:appendix_train}.
\\
To better understand the memory saving of LaMDA, we profile the total memory usage of fine-tuning BART-large on the XSUM dataset for full fine-tuning, LoRA, and LaMDA across different batch sizes. Figure \ref{fig:chart_gpu} shows the peak GPU memory usage for various methods. In specific, LaMDA provides a peak memory saving of up to  $1.32\times$ to fine-tune the BART-large, profiled for different batch-sizes. This saving is primarily due to reduced memory required for activations. Such system-level benefit allows us to fine-tune larger models with larger batch sizes. 

\begin{table}[ht]
\centering
\caption{Comparison of fine-tuning methods for Bart-large. NR denotes not reported. The three values in the last column correspond to R1/R2/RL scores.}
\label{tab:bart-large}
\resizebox{\columnwidth}{!}{
\begin{tabular}{l c c c}
    \toprule
    \textbf{Method}                              & \textbf{\#Params(M)}      & \textbf{XSUM}   & {\textbf{CNN/DailyMail}}       \\
    \cmidrule[\heavyrulewidth]{1-4}
    Full fine-tuning                                        & {415}             & {\textbf{45.14}/\textbf{22.27}/\textbf{37.25}}    & {44.16/21.28/40.90}            \\
    LoRA                                 & {8.6}                & {43.95/20.72/35.68}  & {\textbf{45.03}/21.84/42.15}              \\
    AdaLoRA                           & 8.6                & {44.72}/21.46/36.46       & {45.00/\textbf{21.89}/\textbf{42.16}}         \\
    AFLoRA                                        & {5.1}             & {NR}          & {43.96/21.06/NR}      \\
    LaMDA (LDA-only)                                        & {\textbf{0.20}}             & {40.64/18.11/33.20}          & {40.92/17.53/38.1}      \\
    LaMDA (\(r\)=32)                                        & {{0.85}}             & {43.92/20.68/35.21}          & {44.12/21.16/40.45}      \\
    LaMDA++ (\(r_T\)=32)                                        & {{0.92}}             & {44.32/21.08/36.10}          & {45.01/21.85/42.15}      \\
    \bottomrule

\end{tabular}}
\end{table}

\subsection{Decoder-only Model: LLaMA2}
\label{exp:llama}
We fine-tune and evaluate LLaMA2-7B \cite{DBLP:journals/corr/abs-2307-09288} on complex reasoning task GSM8K \cite{DBLP:journals/corr/abs-2110-14168} and token generative task Wikitext-2 \cite{DBLP:conf/iclr/MerityX0S17} using LaMDA and LaMDA++. The low-rank path is incorporated into the \(\pmb{W_Q}\), \(\pmb{W_K}\), \(\pmb{W_V}\), \(\pmb{W_{FFN1}}\), and \(\pmb{W_{FFN2}}\) matrices in all layers of the model. The hyperparameter \(t_i\) is set to 30\% of the total fine-tuning iterations. For the LoRA and LaMDA experiments, the rank \(r\) is set to 16 and 32, respectively, while the set of potential ranks in LaMDA++ is \(R_S = \{16, 24, 32, 40, 48\}\). We report accuracy for GSM8K and perplexity for Wikitext-2. The results are reported in Table \ref{tab:wiki}; LaMDA and LaMDA++ both surpass LoRA on GSM8K complex reasoning task. And for the Wikitext-2, LaMDA achieves a very close perplexity to that of LoRA, and LaMDA++ outperforms LoRA, while fine-tuning with 5.5$\times$ fewer trainable parameters. This clearly shows the efficacy of LaMDA in yielding improved performance even for complex generative tasks.

\begin{table}[ht]
\centering
\caption{Comparison of fine-tuning results for LLaMA2-7B on GSM8K and Wikitext-2.}
\label{tab:wiki}
\resizebox{\columnwidth}{!}{
\begin{tabular}{l c c c}
    \toprule
    \textbf{Method}                              & \textbf{\#Params(M)}      & \textbf{GSM8K $\uparrow$}   & {\textbf{Wikitext-2 $\downarrow$}}       \\
    \cmidrule[\heavyrulewidth]{1-4}
    LoRA (\(r=16\))                                 & {28}                & {36.9}  & {5.43}              \\
 
    LaMDA  (\(r\)=32)                                      & {\textbf{4.37}}             & {{37.9}}          & {5.45}      \\
    LaMDA++   (\(r_T\)=32)                                     & {{5.12}}             & {\textbf{38.2}}          & {\textbf{5.41}}      \\
    \bottomrule

\end{tabular}}
\end{table}

We also evaluate the performance of LaMDA on commonsense reasoning. We follow the settings in \cite{DBLP:conf/emnlp/HuWLXLB0PL23} and use the Commonsense170K dataset as a combination of training examples of various tasks. Then we evaluate the fine-tuned model on the validation set of each task separately. The collection includes samples of the following datasets: BoolQ \cite{clark2019boolq}, PIQA \cite{DBLP:conf/aaai/BiskZLGC20}, SIQA \cite{DBLP:journals/corr/abs-1904-09728}, the HellaSwag \cite{DBLP:conf/acl/ZellersHBFC19}, WinoGrande \cite{DBLP:conf/aaai/SakaguchiBBC20}, ARC-e and ARC-c \cite{clark2018think}, and OBQA \cite{DBLP:conf/emnlp/MihaylovCKS18}. For this experiment, the set \(R_S\) of LaMDA++ is \(\{32, 48, 64, 80, 96\}\). The fine-tuning results are shown in Table \ref{tab:common}. LaMDA achieves a higher average accuracy than LoRA, while fine-tuning \(\sim11.5\times\) less parameters.

\begin{table*}[ht]
\centering
\caption{Commonsense reasoning results for LLaMA2-7B}
\label{tab:common}
\resizebox{\textwidth}{!}{%
\begin{tabular}{@{}lcccccccccccc@{}}
\toprule
\textbf{Method} & \textbf{\#Params.(M)} & \textbf{BoolQ $\uparrow$} & \textbf{PIQA $\uparrow$} & \textbf{SIQA $\uparrow$} & \textbf{HellaSwag $\uparrow$} & \textbf{WinoGrande $\uparrow$} & \textbf{ARC-e $\uparrow$} & \textbf{ARC-c $\uparrow$} & \textbf{OBQA $\uparrow$} & \textbf{Avg. $\uparrow$} \\ \midrule

LoRA (\(r\)=32)  & 56 & 69.8 & 79.9 & 79.5 & 83.6 & 82.6 & 79.8 & 64.7 & 81.0 & 77.6 \\

LaMDA (\(r\)=64)  & \textbf{4.85} & 71.6 & 80.3 & 79.1 & \textbf{84.0} & 82.4 & \textbf{81.5} & 65.8 & 79.6 & {78.0} \\ 

LaMDA++ (\(r_T\)=64)  & 5.65 & \textbf{71.8} & \textbf{80.6} & \textbf{79.5} & \textbf{84.0} & \textbf{82.7} & \textbf{81.5} & \textbf{66.0} & \textbf{80.6} & \textbf{78.3} \\ 

\bottomrule

\end{tabular}%
}
\end{table*}

\subsection{Ablations and Discussions}
\label{sec:ablation}
\textbf{Impact of initialization choices.} 
A primary step in LaMDA involves initializing the PMA and PMB with the singular vectors (SVs) of the pre-trained weight \(\pmb{W}\). LaMDA utilizes the SVs corresponding to the most significant singular values because, according to SVD theory, these vectors capture the highest proportion of the matrix's total energy-score compared to any other set of \(r\) SVs. Consequently, fine-tuning these vectors has the most significant impact on adaptation. To verify this hypothesis and validate the findings of \cite{meng2024pissa}, we also initialize PMA and PMB with the set of SVs associated with the smallest singular values.

Conversely, VERA \cite{DBLP:journals/corr/abs-2310-11454} initializes the adapters randomly and keeps them frozen. An insightful ablation study would examine the performance of LaMDA when PMA and PMB are initialized randomly, with PMA frozen at the beginning and PMB gradually frozen over time. In this scenario, LDA is initialized to a zero matrix instead of \(I_r\), ensuring that the combined effect of the main path and the adapter path equals the main path at the onset of fine-tuning. 

We fine-tune LLaMA2-7B on GSM8K and Wikitext-2 using the three discussed initialization methods and report the results in Table \ref{tab:init}. For random initialization, we perform Kaiming normal initialization for both PMA and PMB. The remaining hyperparameters are consistent with those in Section \ref{exp:llama}. The Table shows that LaMDA initialized with the first \(r\) SVs outperforms the random initialization when using the same \(r\). Additionally, random initialization surpasses the model initialized with the last \(r\) SVs. The \(r\) is set to 32 for this ablation study. The result underscores the critical impact of fine-tuning the high-energy components of the model.

\begin{table}[ht]
\centering
\caption{Effect of the initialization in LaMDA.}
\label{tab:init}
\resizebox{\columnwidth}{!}{
\begin{tabular}{l c c c}
    \toprule
    \textbf{Initialization}                              & \textbf{\#Params(M)}      & \textbf{GSM8K $\uparrow$}   & {\textbf{Wikitext-2 $\downarrow$}}       \\
    \cmidrule[\heavyrulewidth]{1-4}
    First \(r\) SV                                & {4.37}                & \textbf{37.9}  & \textbf{5.45}              \\
    Last \(r\) SV                                        & {4.37}             & {35.8}         & {5.55}      \\
    Kaiming normal                                        & {4.37}             & {37.1}         & {5.49}      \\
    \bottomrule

\end{tabular}}
\end{table}

\noindent
\textbf{Number of iterations \(t_i\) in gradual freezing.} LaMDA freezes PMB in \(t_i\) first iterations of fine-tuning based on linear schedule. Adjusting this hyperparameter (\(t_i\)) significantly alters the effective number of trainable parameters (\#Params), as the size of PMB (\(r \times d\)) is considerably larger than that of LDA (\(r \times r\)). To investigate the impact of this hyperparameter, we conducted the GSM8K experiment using LLaMA2-7B with various values of \(t_i\). We present the resulting \#Params and accuracy in Table \ref{tab:t_i}. By comparing these results with those in Table \ref{tab:wiki}, we observe that allocating a sufficient number of iterations to training PMB is crucial for surpassing LoRA. Specifically, LaMDA with \(t_i\) set to 10\% of the total iterations fails to outperform LoRA, whereas allocating 20\% and 30\% of the iterations to PMB training results in superior performance relative to LoRA. In the appendix \ref{appn:effective}, we explain how to count the effective number of trainable parameters (\#Params).

\begin{figure}[t]
  \includegraphics[width=0.85\columnwidth]{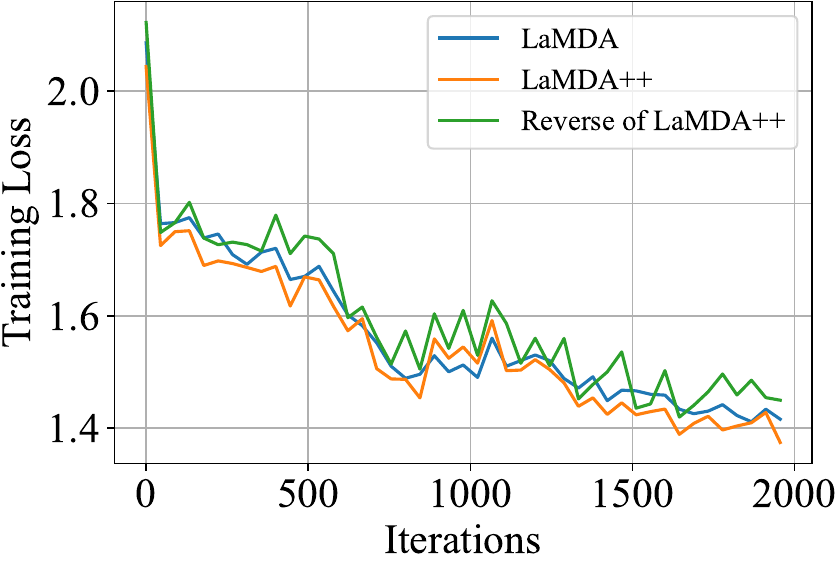}
  \caption{Training Curve of LLaMA2-7B on Wikitext-2.}
  \label{fig:train_convergence}
  \vspace{-4mm}
\end{figure}

\vspace{2mm}
\noindent
\textbf{Effect of the LaMDA++ ranking.} As explained in Section \ref{sec:lamda++}, LaMDA++ generates a sorted list of all linear modules based on the candidacy score \(\nu\). We conduct an essential study to validate the effectiveness of such sorting. First, we allocate ranks according to the list generated by LaMDA++, assigning more ranks to layers with smaller scores. Subsequently, we conduct another experiment where ranks are allocated in the reverse order of LaMDA++, assigning more ranks to layers with higher scores.
The training curves for this experiment are shown in Figure \ref{fig:train_convergence}. The results indicate that LaMDA++ with reverse ordering exhibits noisier training behavior and ends with a higher loss value, translating into higher perplexity on Wikitext-2. Among the three approaches presented in the figure, LaMDA++ demonstrates the lowest training loss, attributable to its appropriate allocation of the rank budget.

\textbf{}

%\section*{Acknowledgments}

% Bibliography entries for the entire Anthology, followed by custom entries
%\bibliography{anthology,custom}
% Custom bibliography entries only
\begin{table}[ht]
\vspace{-4mm}
\centering
\caption{Effect of the fine-tuning iteration \% before freezing PMB.}
\vspace{-3mm}
\label{tab:t_i}
\resizebox{0.7\columnwidth}{!}{
\begin{tabular}{c c c}
    \toprule
    \textbf{\(t_i\)}                              & \textbf{\#Params(M)}      & \textbf{Accuracy $\uparrow$}        \\
    \cmidrule[\heavyrulewidth]{1-3}
    10\% of iterations                                & {1.56}                & 36.1             \\
    20\% of iterations                                       & {2.97}             & {37.0}            \\
    30\% of iterations                                        & {4.37}             & {\textbf{37.9}}            \\
    \bottomrule

\end{tabular}}
\end{table}
\vspace{-6mm}
\section{Conclusion}

In this work, we proposed LaMDA, a novel framework for fine-tuning large language models. LaMDA employs a low-dimensional adapter, significantly reducing the number of trainable parameters and conserving activation memory. The methodology involves freezing the projection matrix \(\pmb{A}\) from the outset and gradually freezing the projection matrix \(\pmb{B}\). We further enhanced LaMDA by incorporating the flexibility of varying ranks across layers, allocating ranks to adapters based on the energy components of the pre-trained weights. Both LaMDA and LaMDA++ demonstrate the capability to facilitate the fine-tuning of larger models on commercial GPUs, offering an efficient and scalable approach to model adaptation.

\newpage
\section{Limitations}
This study has a few limitations. Firstly, the largest model we tested was LLaMA2-7B. Due to time constraints associated with the paper's deadline, we could not extend our experiments to larger models, which could provide further insights into the scalability and effectiveness of LaMDA. Our methodology, LaMDA, has not yet been tested on instruction-following tasks. While the current results are promising, evaluating the performance of LaMDA in these specific tasks is essential to fully understanding its potential and versatility. We plan to address these limitations in future work by conducting experiments on larger models and a broader range of tasks. We are also eager to test the applicability of our method to vision-language models, which was not explored in this paper.
\bibliography{LaMDA}
\pagebreak
\pagebreak

\clearpage
\appendix
\section{Training Details}
\label{sec:appendix_train}
\begin{table}[ht]
\centering
\caption{Hyperparameters for fine-tuning DeBERTa-V3 on GLUE benchmark}
\label{tab:common}
\resizebox{\columnwidth}{!}{%
\begin{tabular}{@{}lcccccccccccc@{}}
\toprule
\textbf{Hyperparameter} & \textbf{CoLA} & \textbf{SST-2} & \textbf{MRPC} & \textbf{QNLI} & \textbf{STS-B} & \textbf{RTE} & \textbf{MNLI} & \textbf{QQP}  \\ \midrule

Learning rate  & 1e-2 & 4e-3 & 8e-2 & 4e-3 & 2e-2 & 4e-2 & 4e-3 & 4e-3 \\

\#Epochs  & 20 & 10 & 20 & 10 & 20 & 20 & 10 & 10 \\ 

Max Seq. Len.  & {512} & {512} & {512} & {512} & {512} & {512} & {512} & {512} \\ 

\bottomrule

\end{tabular}%
}
\end{table}

\begin{table}[ht]
\centering
\caption{Hyperparameters for fine-tuning LLaMA2-7B}
\label{tab:hyper-llama}
\resizebox{0.9\columnwidth}{!}{
\begin{tabular}{l c c c c}
    \toprule
    \textbf{Hyperparameters}                              & \textbf{GSM8K}      & \textbf{Wikitext-2}    & \textbf{Commonsense170k}     \\
    \cmidrule[\heavyrulewidth]{1-4}
    Learning rate                               & {3e-4}                & {3e-4} &{3e-4} \\
    \#Epochs                                        & {6}             & {2} & {3}               \\

    Batch size                                        & {16}             & {16} &{16}               \\
    \bottomrule

\end{tabular}}
\end{table}

\begin{table}[ht]
\centering
\caption{Hyperparameters for fine-tuning BART-large}
\label{tab:hyper-bart}
\resizebox{0.85\columnwidth}{!}{
\begin{tabular}{l c c c}
    \toprule
    \textbf{Hyperparameters}                              & \textbf{XSUM}      & \textbf{CNN/DailyMail}         \\
    \cmidrule[\heavyrulewidth]{1-3}
    Learning rate                               & {2e-4}                & {2e-4} \\
    \#Epochs                                        & {25}             & {15}               \\

    Batch size                                        & {32}             & {64}               \\
    \bottomrule

\end{tabular}}
\end{table}
\noindent
Here we provide the implementation details and the hyperparameters used for training.
In all experiments, we used the PyTorch framework and ADAM \cite{DBLP:journals/corr/KingmaB14} optimizer. 
\subsection{DeBERTa-V3}
To fine-tune DeBERTa-V3 on the GLUE benchmark, we use a batch size of 32 and use the following setup for the learning rate and number of epochs, which are similar to what \cite{DBLP:journals/corr/abs-2310-08659} used.

\subsection{BART-large}
For fine-tuning BART-large on XSUM and CNN/DailyMail we set the maximum input sequence to 1024 and the maximum target sequence to 128. Learning rate, number of epochs, and batch size are shown in the Table \ref{tab:hyper-bart}, which are similar to what \cite{DBLP:journals/corr/abs-2310-08659} used.

\subsection{LLaMA2-7B}
We follow the setting of \cite{DBLP:journals/corr/abs-2310-08659} to fine-tune LLaMA2-7B on GSM8K and Wikitext-2 datasets. Moreover, we adopt the Commonsense170K dataset from \cite{DBLP:conf/emnlp/HuWLXLB0PL23} and use the default setup to fine-tune LLaMA2-7B for commonsense reasoning. For evaluation, we use lm-evaluation-harness library \cite{eval-harness}. The hyperparameters are provided in Table \ref{tab:hyper-llama}.

\section{Effective number of trainable parameters (\#Params) in LaMDA}
\label{appn:effective}
Assuming \(L\) trainable linear modules in the model, \(t_i\) initial iteration for gradual freezing, and \(T\) total iterations, the effective number of trainable parameters can be computed as below:
\begin{equation}
    \#\mathrm{Params}=\sum_{l=1}^{L}[\frac{t_i}{T}\times \frac{\mathrm{NP}(PMB_l)}{2} + \mathrm{NP}(LDA_l)]
\end{equation}
where \(\mathrm{NP}(\pmb{X})\) is a function that counts the number of trainable elements in the matrix \(\pmb{X}\); \(PMB_l\) and \(LDA_l\) are the projection matrix B and low-dimensional adapter in the linear module \(l\).
\end{document}